# Contextual Relationship-based Activity Segmentation on an Event Stream in the IoT Environment with Multi-user Activities

Minkyoung Cho, Younggi Kim, Younghee Lee
Korea Advanced Institute of Science and Technology
cho_omk@kaist.ac.kr, younggi521@kaist.ac.kr, yhlee@cs.kaist.ac.kr

## ABSTRACT
The human activity recognition in the IoT environment plays the central role in the ambient assisted living, where the human activities can be represented as a concatenated event stream generated from various smart objects. From the concatenated event stream, each activity should be distinguished separately for the human activity recognition to provide services that users may need. In this regard, accurately segmenting the entire stream at the precise boundary of each activity is indispensable high priority task to realize the activity recognition. Multiple human activities in an IoT environment generate varying event stream patterns, and the unpredictability of these patterns makes them include redundant or missing events. In dealing with this complex segmentation problem, we figured out that the dynamic and confusing patterns cause major problems due to: *inclusive event stream, redundant events, and shared events*. To address these problems, we exploited the contextual relationships associated with the activity status about either ongoing or terminated/started. To discover the intrinsic relationships between the events in a stream, we utilized the LSTM model by rendering it for the activity segmentation. Then, the inferred boundaries were revised by our validation algorithm for a bit shifted boundaries. Our experiments show the surprising result of high accuracy above 95%, on our own testbed with various smart objects. This is superior to the prior works that even do not assume the environment with multi-user activities, where their accuracies are slightly above 80% in their test environment. It proves that our work is feasible enough to be applied in the IoT environment.

## 1. INTRODUCTION
The era of the internet of things (IoT) opens a new horizon to enhance the quality of human life, via highly advanced technologies that realize the Ambient Assistant Living (AAL). To achieve the AAL, smart objects are becoming more prevalent in our environment, and technologies that can support the necessary services in the specific environment are required. Specifically, the human activity recognition (AR) recognizes the actions and goals of humans from observing the human behaviors and the contextual conditions. So, the activity recognition constitutes the underlying technology for the AAL in the IoT environment where many unpredictable behaviors by individual participants are found.

Accordingly, numerous researches have been proposed to apply the AR to the real world. [11, 12] claimed that the nature of human activities in the IoT environment poses the challenges of *concurrent activities, interleaved activities, ambiguity, variety, and multiple subjects*, all of which complicate the issue of AR. In general, the AR involves 1) monitoring and collecting event streams, 2) activity segmentation, and 3) activity recognition.

In this paper, the events are defined as the data generated from the deployed state-change sensors or smart devices upon detecting the human behaviors. Examples of the events include *entrance*, *sit down* or *light on*. Thus, an activity such as a *seminar, study* or *phone call* is expressed as an event stream generated by the smart objects actuated from that activity. In our environment, the event streams from a series of activities are concatenated into one longer event stream. Therefore, the long event stream must be segmented in advance according to genuine activity boundaries for precise activity recognition, which we call activity segmentation.

For the activity recognition, various researches have been proposed. Earlier works focused not on activity segmentation but on activity recognition from a pre-segmented event stream. [14, 18, 20] assumed pre-segmented event streams provided by people, where they recorded the starting and finishing times for every conducted activity. However, such manual segmentation is physically demanding, time-consuming, and error-prone so that their assumption may not be practical. Other studies [1, 2, 10, 17] conducted activity segmentation using the concept of a window based on the time or the number of events. But, the proper length of the window is difficult to determine because the duration or the length of event stream of each activity may differ significantly. Furthermore, in a case of the consecutive windows, if the length of the window is incorrect, the accuracy of activity recognition would be gradually decreased due to cumulative deviation from the genuine activity boundaries.

To address the activity segmentation problem, different approaches [3, 13, 15, 16, 17, 21] have been proposed. Most of them focused on a smart home environment, where the activities can be distinguished by the situation information only. The situation information includes the occurrence time and/or location, the location-specific objects, or the predefined event set for each activity. Unlike these studies, we assumed the IoT environment where the location-specific and the time-specific situation information are not available, and various activities using almost the same objects can occur in succession.

In this IoT environment, the event streams generated from the various activities may show dynamic and confusing patterns so that it is necessary to elaborately analyze them and grasp the contextual relationships inherent in a series of event streams. In other words, we should be able to determine segment boundaries through capturing the contextual relationships associated with the activity status about ongoing or terminated/started. To figure out these relationships, we took advantage of the Long short-term memory (LSTM) model [9], which has long been used in the text fields [4–6]. Since the LSTM model can ascertain the contextual relationships between sufficient length of component events in the dynamic and confusing pattern sequence, we utilized the model to determine the segment boundaries in the concatenated long event stream. Based on its learning, the LSTM model can infer and determine whether or not an activity is terminated, and outputs the boundary information in case of terminated status.

To apply the LSTM model for the activity segmentation, we designed our specialized input and target vector weight. In addition, we devised a time interval validation algorithm for enhancing the result from the LSTM-based method. Our approach resulted in the surprising accuracy of 96.77% in our testbed and proved its feasibility in the real world.

The rest of this paper is organized as follows. In Section 2, we analyze the event streams from the activities and define major problems in the activity segmentation. In Section 3, we present our LSTM method and propose a lightweight yet efficient validation algorithm. In Section 4, we evaluate our approach in our testbed for the IoT environment and demonstrate our work. After discussing related work in Section 5, we conclude in Section 6.

## 2. ANALYSIS AND PROBLEMS

The event streams generated from various activities occurring in the IoT environment may show dynamic and confusing patterns. In this regard, we analyzed the event streams collected from our testbed over a course of nearly nine months and discovered the following characteristics inherent in the event streams.

**1) Streams with similar component events from different activities:** Even if the activities differ, the event streams from the activities would be similar if they occur in the same space. State-change sensors in an IoT environment are related not only to one activity but to several activities in that space, so their event streams would have a similar set of component events. Figure 1(a) shows the probability that each event is included in the major activities.

**2) Different event streams from an identical activity**: As opposed to 1), event streams from an activity may be composed of different events according to the situation such as the weather, the number of participants, and the occurrence time. From the '*study*' activity, the following three event streams can be generated.

1. entrance – light on – sit down – stand up – light off – exit
2. entrance – sit down – stand up – exit
3. entrance – air conditioner on – light on – sit down – stand up – light off – air conditioner off – exit

The first stream can occur at night and the second one in the morning. When the activity is conducted in summer, the stream may resemble the last stream.

**3) Event stream for an undefined activity**: For segmentation, the set of activities should be defined as many as possible in advance. However, in the IoT environment, undefined activities may occasionally happen. For example, someone can get into the environment to look for missing stuff or other people. In this case, event streams composed of '*entrance*' and '*exit*' are generated from the undefined activities. In this paper, we do not consider the occurrence of the undefined activities, since the portion of them is almost negligible in our environment.

**4) Interleaved events stream of simultaneous activities**: If multiple people exist in the same environment, more than one activity can occur simultaneously. For example, while one user is studying, another user can receive a phone call. In this case, simultaneous event streams for the activities would be interleaved, making the activity segmentation much more difficult. However, when multiple activities occur in a space, they can disturb or distract with each other so that such situation would rarely happen in general. For this reason, we do not consider this situation, and this issue remains as our future work along with the above case.

These characteristics cause the major problems that make the activity segmentation problem complicated to deal with.

**Inclusive event stream problem**. A pattern, which is very similar to one event stream of one activity, can be found within an event stream from other longer activity. This problem may occur between the event stream of a simple activity lasting a short time by a single user and that of a complex one lasting a long time by multiple users.

*Activity α: entrance − sit down − stand up − exit*

From a simple activity, such as '*phone call*', its complete event stream can be illustrated as above. However, from a complex one, such as '*seminar*', the above pattern can be included within the entire event stream from the activity. So, a complete event stream is likely to be erroneously segmented within the segment body.

**Redundant events problem**. When an activity is conducted by a single user or multiple users, the event stream may include repetitive events due to unpredictable or emergent situations by the individual users as follows:

*Activity α: entrance − light on − **sit down − stand up**
− exit − entrance − **sit down − stand up** − light off − exit*

This event stream that has the repetitive events is one instance for the '*study*' activity when someone leaves the space to go to the bathroom. In addition, if more than one person participates the activity, the event stream would be longer with redundant events. This is because the same events such as the '*exit*' can be generated by all the individuals and the occurrence time of each event may be different. Figure 1(b) shows the average probability for each event to occur at a certain part in an event stream. Here, when we divide each event stream into three parts of equal length, then the first part

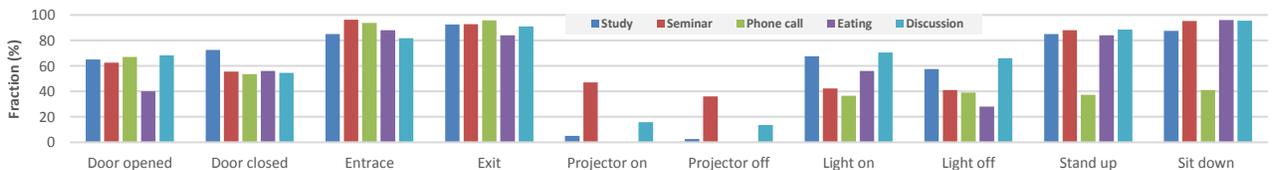
**(a) Probability of each event included in the major activities**

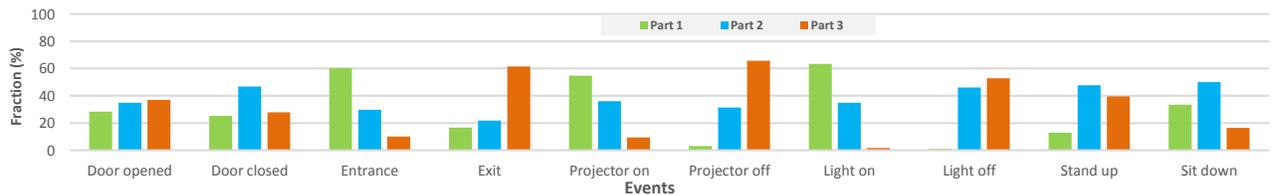
**(b) Probability of each event found in one of the three even parts of an event stream**
**Figure 1. Characteristics of events generated in the IoT environment**

will include the starting point and the third part will include the finishing point. However, those boundary-specific events may occur in any other parts. Therefore, since the events only do not guarantee the starting or finishing contexts, determining the segment boundaries depending on these events only is not feasible.

**Shared events problem over consecutive activities**. When an activity occurs immediately after a previous activity, a set of events which are commonly related to the two activities is not generated in general. As one instance, a user can receive a phone call while studying. To conduct the *'phone call'*, the user does not have to actuate the light and sit down again, already actuated from the *'study'*. In this case, the event stream is illustrated as below.

*Activity α: entrance − light on − sit down*

*Activity β: stand up – exit*

Therefore, shared events are rarely generated from the subsequent activity of two consecutive activities, and the segment boundary would be very ambiguous in the concatenated event stream.

**Interleaved segment boundary problem**. When more than one activity occurs simultaneously, the boundary of one activity tends to be interleaved with the events of other activities. This effect will make the segment boundaries ambiguous to determine.

## 3. ACTIVITY SEGMENTATION METHOD

If a noisy input sequence has long time steps, [6] claims that the LSTM model can perform well than the Hidden Markov models (HMMs) and the standard Recurrent Neural Networks (RNNs) as sequence processing methodologies. Based on this feature, we adopt and apply the LSTM model to the activity segmentation by rendering it to capture the contextual relationships inherent in a long series of events. Our target method has been developed in a few stages. At the first stage, we customized the LSTM model into three layers of equal length, where we found the appropriate time steps in the layer and addressed the bias problem. Then, each LSTM input was augmented with some object status information and solved the *inclusive event stream problem* using an optimal set of the object status. Lastly, the segmented boundaries from the previous stage were inspected with our time interval validation algorithm so that the final result could be enhanced. All these stages are elaborated in the following subsections.

## 3.1 LSTM Model for Activity Segmentation

### 3.1.1 Basic LSTM Model

Our LSTM model is composed of three layers, which are input layer, hidden layer, and output layer, where each layer has an identical length of nodes known as time steps. Figure 2 shows our LSTM model where an input sequence with the length of time steps enters into the input layer and then the hidden layer passes its calculated results to the output layer. In our input procedure, an input sequence enters into the model, shifting one event in the long concatenated event stream for each input sequence. In one input sequence, each event enters into its matching input node after converted into the one-hot vector. Each output node has a two-dimensional vector, where if the first element of the vector is larger than the second one, the event for that output node is inferred as the segment boundary. Therefore, we can get all the indices of events in the long event stream, inferred as the segment boundaries.

In our model, we set the length of time steps to 60, because more than 99% of the lengths of all the event streams are less than 60. We can see the distribution of the lengths in Figure 3. In the distribution, another noticeable factor is that the minimum length of the conforming activity is found to be 4 in our testbed.

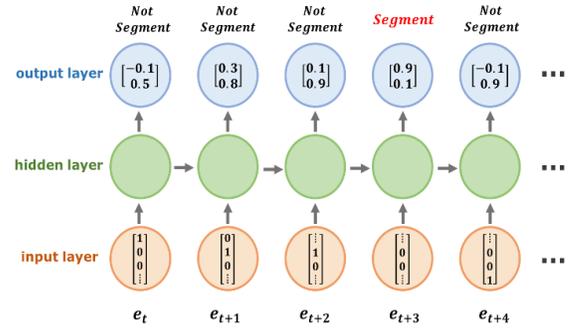

**Figure 2. LSTM model for activity segmentation**

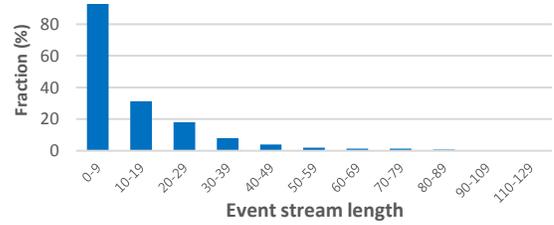

**Figure 3. Distribution of the lengths of event streams**

In the training process of the LSTM, our cost value is defined as the sum of squares of the differences between the two-dimensional vectors in each output node and its matching target vector, which is the typical definition in this area. The training process is executed iteratively to attain the lower cost value and it works better when the data are well-distributed over multiple groups. However, the dataset for the activity segmentation consists of only two groups, one for the segment boundaries and the other for the segment bodies. Here, there is a large unbalance between the numbers of events in two groups, which may lead to a bias problem in the training process. That is, the training process using two severely unbalanced numbers of data groups would decrease the segmentation accuracy.

To address this bias problem, it is necessary to counterbalance the biased effect. So, the target vector is weighted by the value of ω which is close to the degree of bias, as in the following formulation.

$$\text{weight } \omega = \sqrt{\frac{(e_{total} - e_{boundary})}{e_{boundary}}},$$

where $e_{boundary}$ denotes the number of segment boundaries and $e_{total}$ is the total number of events, so the square root indicates the degree of bias. In the evaluation, we determined and validated the optimal weight ω empirically.

### 3.1.2 LSTM with Object Status

The basic LSTM model described above results in the accuracy of 75% during the activity segmentation process in our testbed. We found in our experiments that the basic model rarely solves the *inclusive event stream problem*. Therefore, if the patterns trained as short and complete event streams, such as *entrance – sit down – stand up- exit,* are observed in an input sequence, the model tends to regard each pattern as a single complete event stream, even if the stream is just a part of true event stream from one activity.

To address the *inclusive event stream problem* frequently observed in the basic model, we investigated various situations around the segment boundaries and we found the following property:

*An event that may suggest the boundary can have different contextual meaning from the same event occurring at other position in one activity.*

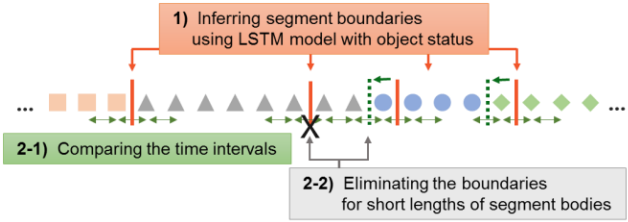

**Figure 4. Overall activity segmentation procedure**

As to this property, the contextual meaning represents the progress of the activity, and the different meanings of the same event can be derived from the interrelationships with its neighboring events. Since such neighboring events are generated by the objects, it is meaningful to reflect more elaborate object status into the LSTM model. So, we utilized major object status to determine whether or not the event is relevant to the segment boundary.

In detail, we designed augmented input vector of the LSTM model by appending the additional object status to the original one-hot vector. We selected the additional objects based on our criterion that the target object should be relevant to more than 90% of all event streams in our testbed. The Table 1 presents a list of the selected objects and their relevant events. In appending the object status to the input vector, the type of the object status is considered. For the object with on or off status, we append 1 or 0 respectively. In cases of the people count or occupied seats count, we append the value to the input vector after dividing the value by the maximum value for normalization. In applying this scheme, we considered not only a single object but also combined objects in augmenting the input vector. Based on various cases with these objects, we found out the target objects that resulted in the optimal performance. These procedures in the first stage are shown as step 1) in Figure 4.

## 3.2 Time Interval Validation Algorithm

If a prevailing event suggesting a segment boundary, which can occur at any position in an event stream as in Figure 1 (b), is generated before or after the true boundary, it can make a bit shifted output boundary. To prevent this shifted boundary, we can exploit another contextual property associated with the time intervals between consecutive events. In general, segmenting the event stream by using mainly the time intervals, which was used by some of the existing approaches, may not be practical in the IoT environment. This is because time interval-based segmentation can be incorrect in the cases of consecutive activities without a threshold time interval, or consecutive events with a long time interval in an activity. Even though the time intervals cannot be used solely for the activity segmentation, we observed a meaningful property pertaining to them. That is, the time intervals between consecutive events, around the starting or finishing points in each activity, tend to be small, compared to the intervals between neighboring activities. Specifically, when an activity is about to terminate, a set of events such as *stand up, door opened, and exit* is generated consecutively within a relatively short time. We can also observe this kind of property when an activity has just started.

So, we intended to utilize this property for enhancing the accuracy of our LSTM method. The simple way to utilize this property is comparing the time intervals around the inferred segment boundaries. Thus, we propose our lightweight yet efficient enhancing algorithm that can examine the validity of the segment boundaries and confirm them. This algorithm works in two steps. The first step calculates three time intervals around each inferred segment boundary. The first interval is calculated between the last event and the last but one in the preceding segment body before the segment boundary. The second interval is computed at the segment boundary, which is between the last event in the previous segment and the first event in the subsequent segment. In that subsequent segment, the third time interval is obtained between the first and the second events. Then, we select the largest interval and determine its later event as the revised segment boundary. The second step confirms whether or not the lengths of all the segments exceed the minimum length of an event stream. If any segment has a length less than the minimum length, then we compare the two time intervals toward the two segments which are before and after that segment, and eliminate the segment boundary corresponding to the smaller interval. This algorithm is simple yet very efficient, which is demonstrated in the evaluation section.

**Table 1. Selected objects and their relevant events**

| Object | Relevant Events |
|---|---|
| People count | Entrance, Exit |
| Light | Light on, Light off |
| Door | Door opened, Door closed |
| Occupied seats count | Sit down, Stand up |

## 4. EVALUATION

To demonstrate our approach, we used our own IoT environment installed in a room in our campus (building N1, KAIST). The environment has 12 state-change sensors such as *Door sensor, Light sensor, Presence sensor, and Seat sensor*. In the space, 23 different events are generated from these objects and 17 different activities are defined and assumed to occur. Our dataset was collected from September 2015 to June 2016. The concatenated event stream used in the experiments and evaluations is composed of 6,843 events generated from 436 activities, which is obtained through preparation. In the preparation, the event stream was segmented manually for learning the true segment boundaries and evaluating our approach. And if the original raw data contained some erroneous event streams, like that the duration of a stream is too short to be a defined activity or abnormally arranged events are generated, they are regarded as undefined or nonconforming activities and removed in the evaluation.

To implement our LSTM model, we utilized TensorFlow [19] which is an open source software library for machine learning. In the evaluation, we utilized 90% of the events for training and the remaining events for testing the model. We set the length of time steps to 60 and each input sequence of the LSTM is generated by sliding one event in both training and testing processes.

The main metrics for the performance evaluation are three metrics: the recall, precision, and F1-score. In our experiments, the recall indicates the ratio of the number of correct segment boundaries that we found to the number of the true segment boundaries. The precision is the ratio of the number of correct segment boundaries among the number of the segment boundaries that we found.

## 4.1 Target Vector Weight for Bias Problem

In the LSTM model, it is necessary to consider the bias problem that occurs when the number of events relevant to the segment boundaries is much smaller than the number of events for the segment bodies. Since we set the length of time steps to 60 and given that the length of the shortest event stream of an activity is 4, then the number of segment boundaries is between 1 and 15 per time steps. In practice, only 435 out of 6,843 events are relevant to the segment boundaries, with the remaining 6,572 events included in the segment body in our testbed. Thus, in the training process, the target vector for the segment boundary is weighted according to the formulation defined in Section 3. We determined the weight to be 3 which achieves the highest accuracy as shown in Figure 5.

**Table 2. Baseline vs. our LSTM methods**

| Method | recall (%) | precision (%) | F1-score (%) |
|---|---|---|---|
| siHMM | 46.36 | 45.95 | 46.15 |
| Basic LSTM | 89.36 | 64.62 | 75.00 |
| Basic LSTM + Time interval algorithm | 89.36 | 73.68 | 80.77 |

**Table 3. Performance of the LSTM with object status**

| Object | recall (%) | precision (%) | F1-score (%) |
|---|---|---|---|
| Basic LSTM | 89.36 | 64.62 | 75.00 |
| People count | 93.62 | 77.19 | 84.62 |
| Light | 93.62 | 80 | 86.27 |
| Door | 82.98 | 70.91 | 76.47 |
| Occupied seats count | 91.49 | 93.48 | 92.47 |
| **People count + Light** | **93.62** | **95.65** | **94.62** |
| People count + Door | 93.62 | 77.19 | 84.62 |
| People count + Occupied seats count | 95.74 | 84.91 | 90.00 |

### 4.2 Baseline vs. Our LSTM Methods
This paper utilizes the siHMM [15] as the baseline method for comparison. The siHMM is an outstanding HMM for the activity segmentation with state-of-the-art accuracy compared to other HMMs. Table 2 shows the recall, precision and F1-score resulted from the siHMM and our LSTM methods, which were conducted using our dataset. From these results, our proposed methods show much higher accuracy. The accuracies can be interpreted that the large difference is caused by the attributes of our dataset, where the events in an event stream are dependent on each other. While the HMM-based approach is inappropriate when events are inter-dependent, the LSTM can discover the intrinsic sequential relationship between events within a segment body or around the segment boundary. Therefore, our LSTM-based approach is suitable for the IoT environment.

It is also important to note the precision values from the basic model and the time interval validation algorithm enhancing the result. When we applied the algorithm to the resulting segment boundaries, a bit shifted boundaries described in subsection 3.2 were revised to the correct boundaries, so the precision increased by 9% and the final F1-socre increased by 5.77% to 80.77%. Thus, we could get higher accuracy by our validation algorithm.

### 4.3 LSTM with Object Status
In examining the detailed results, we found that the LSTM did not determine frequently the segment boundaries in the case of the *inclusive event stream problem*. In other words, the basic model regarded a part of a long event stream from one complex activity as a short and complete event stream from the simple activities, such as the *'phone call'* activity. This problem made the precision decreased compared to the high recall. To address the problem, we augmented each input vector of the LSTM with some prevailing object status information that meets our criterion in Section 3. Table 3 shows the results of the experiments with the cases of appending one object or combined objects. As shown in the table, when an object status or a set of object status are appended to the original input vector, the precision values show variations, while most recall values increase meaningfully, causing the F1-score to be enhanced in general. These results suggest that the appended object status can help the LSTM to resolve *the inclusive event stream problem* by grasping more insightfully the contextual meaning of each component event by the property described in subsection 3.1.2.

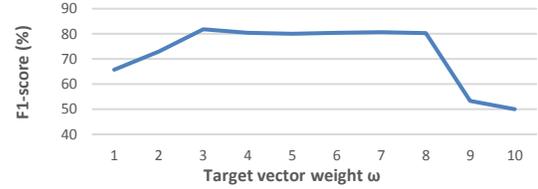

**Figure 5. F1-score according to target vector weights**

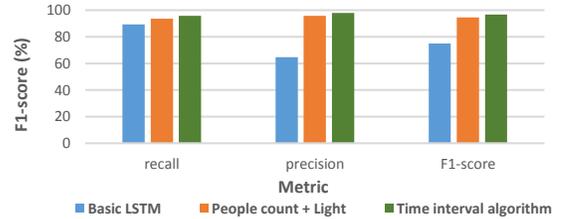

**Figure 6. Basic LSTM vs. LSTM with object status vs. Time interval algorithm**

While the concept of the object status is useful for the activity segmentation, using the concept solely for the process cannot solve the major problems declared in Section 2. For example, if the segment boundary is determined by the people count only, unpredictable *'entrance'* or *'exit'* event by some users can easily lead to wrong segmentation. Thus, the object status should be used adjunctively with the event stream due to the low flexibility and ambiguity in applying the criterion for segmentation.

From the results of activity segmentation by appending the object status into the basic LSTM, we found that using the *people count and light* achieved the highest accuracy of 94.62%. This result can be interpreted that since the light and people count generate the dominant events that can signify the segment boundary, their combination would yield the optimal performance. This analysis can be strongly supported by the Figure 1(b). This result was obtained without applying our time interval validation algorithm.

As our final stage evaluation, we applied the proposed validation algorithm to the above result. The final result showed the surprising accuracy level of 96.77%, by correcting a bit shifted segment boundaries and eliminating non-boundary segmentation. Thus, we demonstrated that this algorithm could enhance the accuracy even further from the high accuracy of the previous results. Figure 6 shows this enhancement.

## 5. RELATED WORK
The recent study [15] is closest to our study in that the authors attempted to use mainly the event streams by proposing the siHMM, which is a feasible model for data stream segmentation in their environment. The most important problem of the original HMMs is that they cannot distinguish the dynamics between the inter-segment and intra-segment. The problem was addressed by the siHMM, which learns various data streams and grasps the sequential relationships to determine the segment boundaries like our LSTM model. However, an event stream generated from the IoT environment has intrinsic order between its component events, so events are not independent of each other. Despite the fact that using the siHMM results in state-of-the-art accuracy in their dataset, the model can have low performance in the IoT environment. When we applied the siHMM to our testbed, the resulting accuracy was less than 50%, i.e., much lower than our LSTM method.

## 6. CONCLUSION AND FUTURE WORK

The IoT environment requires highly advanced activity recognition technology to provide proper services to multiple participants. Activity recognition technologies have advanced significantly, and they assumed that an event stream used for the recognition is given by the unit of one activity. So, a concatenated event stream that is generated from a series of multi-user activities should be segmented in advance according to the activities. In this paper, this activity segmentation problem was addressed, which is difficult because unpredictable actions by the individuals make the event stream patterns dynamic and confusing.

In our contributions, we found important characteristics of the event streams and figured out major problems for activity segmentation: *the inclusive event stream, the redundant events, and the shared events problems.* To determine the segment boundaries in a concatenated event stream, we exploited the contextual relationships about the activity progress, which could be captured within the segment body and around the segment boundaries. For this purpose, the LSTM model was rendered for our environment via devising its input vector and target vector weight, and its resulting boundaries were validated by our lightweight yet efficient algorithm. Through these experiments, our study showed the surprising segmentation accuracy in our testbed and demonstrated its feasibility in the IoT environment.

In the future work, we will consider the remaining possible situations where undefined activities may occur and a few multi-user activities occur simultaneously. By developing our activity segmentation methodology, we will present more feasible and optimal approach for the IoT environment.

## 7. ACKNOWLEDGMENTS

This work was supported by ICT R&D program of MSIP/IITP [B0101-16-0334, Development of IoT-based Trustworthy and Smart Home Community Framework].